\documentclass{article}
\usepackage{icml2002}
\usepackage{epsf}
\usepackage{mlapa}

\usepackage[dvips]{graphicx}
\usepackage{latexsym}
\usepackage{xspace}
\usepackage{multirow}
\usepackage{amsmath}
\usepackage{amsthm}
\usepackage{amssymb}
\usepackage{color}
\usepackage{algorithm}
\usepackage{algorithmic}
\usepackage{url}
\usepackage{subfigure}

\newcommand{\BE}{\begin{enumerate}}
\newcommand{\EE}{\end{enumerate}}
\newcommand{\BI}{\begin{itemize}}
\newcommand{\EI}{\end{itemize}}

\definecolor{gray}{rgb}{0.5,0.5,0.5}

\newcommand{\RE}{\mathbb{R}}
\newcommand{\E}{\mathbb{E}}
\providecommand{\nb}{\nabla}

\newcommand{\cG}{\mathcal{G}}

\newcommand{\cM}{\mathcal{M}}

\newcommand{\cS}{\mathcal{S}}
\newcommand{\cU}{\mathcal{U}}

\newcommand{\cY}{\mathcal{Y}}

\newcommand{\Bmu}{\bar{\mu}}

\newtheorem{assumption}{Assumption}

\newtheorem{theorem}{Theorem}

\newcommand{\OMEGA}{\omega(h | \phi, g, y)}
\newcommand{\OMEGAT}{\omega(g_{t+1} | \phi, g_t, y_t)}

\newcommand{\NU}{\nu(y|i)}

\newcommand{\MU}{\mu(u | \theta, h, y)}
\newcommand{\MUT}{\mu(u_t | \theta, g_{t+1}, y_t)}

\newcommand{\QJIU}{q(j | i, u)}

\newcommand{\NI}{n_\phi}
\newcommand{\NP}{n_\theta}
\newcommand{\NS}{|\cS|}
\newcommand{\NO}{|\cY|}
\newcommand{\NY}{\NO}
\newcommand{\NA}{|\cU|}
\newcommand{\NB}{|\cG|}
\renewcommand{\NG}{\NB}

\providecommand{\GPOMDP}{\textsf{GPOMDP}}

\newcommand{\IGPOMDP}{\textsf{IState-GPOMDP}}

\newcommand{\EGPOMDP}{\textsf{Exp-GPOMDP}}
\newcommand{\GAMP}{\textsf{GAMP}}
\newcommand{\WILLR}{Williams \textsf{REINFORCE}}

\providecommand{\TD}{\textsf{TD($\lambda$)}}

\newcommand{\BELIEF}{\textsf{Belief}}

\begin{document} 

\twocolumn[
\icmltitle{Scaling Internal-State Policy-Gradient Methods for POMDPs}
 
\icmlauthor{Douglas Aberdeen}{douglas.aberdeen@anu.edu.au}
\icmladdress{Research School of Information Science and Engineering, 
             Australian Nat. University, ACT 0200, Australia}
\icmlauthor{Jonathan Baxter}{jbaxter@panscient.com}
\icmladdress{Panscient Pty Ltd, Adelaide, Australia}
\vskip 0.3in
]

\begin{abstract}
  Policy-gradient methods have received increased attention recently
  as a mechanism for learning to act in partially observable
  environments. They have shown promise for problems admitting
  memoryless policies but have been less successful when memory is
  required.  In this paper we develop several improved algorithms for
  learning policies with memory in an infinite-horizon setting ---
  directly when a known model of the environment is available, and via
  simulation otherwise.  We compare these algorithms on some large
  POMDPs, including noisy robot navigation and multi-agent problems.
\end{abstract}
\section{Introduction}

Partially observable Markov decision processes (POMDPs) provide a
framework for agents that learn how to interact with their environment
in the presence of multiple forms of uncertainty, particularly
observation noise. The only performance feedback given to the agents
is a scalar reward signal. The rewards can be noisy and delayed from
the actions that caused them. The agent's goal is to learn a policy
that maximises the long-term average reward. 

Unlike the situation with fully observable MDPs, the problem of
finding optimal policies for POMDPs is PSPACE-complete
\cite{papadimitriou:1987}, even when the dynamics are known. Thus,
approximate methods are required even for relatively small problems
with known dynamics. There are two main classes of approximate
algorithms: value-function methods that seek to approximate the value
of probability distributions over the state space or \emph{belief
  states} (see \npcite{hauskrecht:2000}, for a good overview); and
policy-based methods that search for a good policy within some
restricted class of policies. In this paper we introduce three new
approximate algorithms of the latter variety.

All three algorithms use finite state controllers (FSCs) to store
information about the history of observations. Both the
transitions between the internal states of the controller given an
observation, and the distribution over the agent's actions given a
particular internal state, are governed by parameterised classes of
functions. All three algorithms are {\em policy-gradient} algorithms
in the sense that the agent adjusts the function parameters in the
direction of the gradient of the long-term average reward. The
algorithms differ in the way the gradient is computed. The first
algorithm---\GAMP---relies on a known model of the POMDP to compute
the gradient. It is based on a series matrix expansion of an exact
expression for the gradient.  The second
algorithm---\IGPOMDP---computes the gradient stochastically by
sampling from the POMDP and the internal-state trajectories of the
FSC.  The third algorithm---\EGPOMDP---reduces the variance of
\IGPOMDP\ by computing true expectations over internal-state
trajectories.

We give convergence theorems and present experimental results for
several different problems including a multi-agent task with 21,632
states requiring cooperation for its solution. We have omitted proofs
for brevity: they may be found in the technical report
\singleemcite{aberdeen:2002}.




\section{POMDPs with Internal-State}
\label{sec:part-observ-mark}

Our setting is that of an agent taking actions in a world according to
a parameterised policy. The agent seeks to adjust the policy in order
to maximise the long-term average reward.  A natural model for this
problem is that of POMDPs.  For ease of exposition we consider finite
POMDPs,\footnote{Extensions to infinite POMDPs are relatively
straightforward, but add significant mathematical complexity without
fundamentally altering the model.} consisting of: states
$\cS=\{1,\dots, \NS\}$ of the world; actions $\cU =\{1,\dots,\NA\}$
available to the policy in each state; observations $\cY =\{1,\dots,
\NY\}$ and a (possibly stochastic) reward $r(i)$ for each state $i\in
\cS$.

Each action $u\in\cU$ determines a stochastic matrix $Q(u) = [ \QJIU
]$ where $\QJIU$ denotes the probability of making a transition from
state $i\in \cS$ to state $j\in\cS$ given action $u\in\cU$.  For each
state $i$, an observation $y\in\cY$ is generated independently with
probability $\nu(y|i)$.

\begin{figure}
\begin{center}
\input{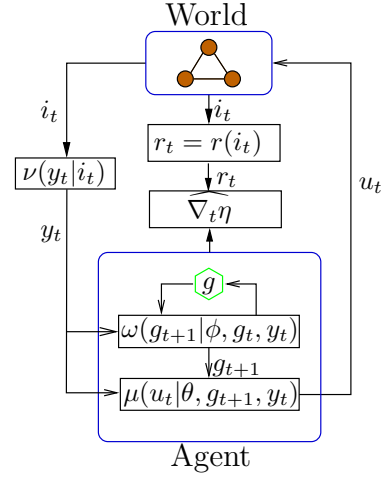}
\caption{Evolution of a finite state POMDP controller.} 
\label{f:fsc-pomdp}
\end{center}
\end{figure}

The policy has a finite set---$\cG = \{1,\dots,\NB\}$---of internal
states, or I-states, with which to store information about the
observation history. The policy is stochastic with the probability of
choosing action $u$ given observation $y$, I-state $g$, and parameters
$\theta\in\RE^{\NP}$, written as $\mu(u|\theta,g,y)$.  The I-state
itself evolves stochastically as a function of the current observation
and a second set of parameters $\phi\in\RE^{\NI}$.  The probability of
making a transition from I-state $g\in\cG$ to I-state $h\in\cG$ is
$\omega(h | \phi,g, y)$.  With these definitions the agent may be
viewed as a stochastic {\em finite state controller} or FSC. The
evolution of world/I-state pairs $(i,g)$ (Figure~\ref{f:fsc-pomdp}) is
Markov, with an $\NS\NB\times \NS\NB$ transition probability matrix
$P(\theta,\phi) = [p(j,h|\theta,\phi,i,g)]$ whose entries are given by
\begin{equation}
\label{eq:bigp}
p(j,h|\theta,\phi,i,g) = \sum_{y,u} \NU
\OMEGA \MU \QJIU. 
\end{equation}
\begin{assumption}
\label{ass:station}
Each $P(\theta,\phi)$ has a unique stationary distribution
$\pi(\theta,\phi) := [ \pi_{1,1}(\theta, \phi), \dots,
\pi_{\NS,\NB}(\theta,\phi)]'$ satisfying 
$\pi'(\theta,\phi) P(\theta,\phi) = \pi'(\theta,\phi)$.
\end{assumption}
\begin{assumption}
\label{ass:bound}
We assume the magnitudes of the rewards, $|r(i)|$, are
uniformly bounded by $R < \infty,~\forall i$.
\end{assumption}
\begin{assumption}
\label{ass:deriv}
The derivatives,
$\frac{\partial \MU}{\partial \theta_k}$
and $\frac{\partial \OMEGA}{\partial \phi_l}$
are uniformly bounded by $U < \infty$ and $Q < \infty$ respectively, $\forall g,h\in\cG$, $u\in \cU$, $y\in \cY$, $\theta \in \RE^{\NP}$
and $\phi\in\RE^{\NI}$.
\end{assumption}
\begin{assumption}
\label{ass:ratiobound}
The ratios 
$$
\frac{\left|\frac{\partial \MU}{\partial
\theta_k}\right|}{\MU}
\quad \text{and} \quad
\frac{\left|\frac{\partial \OMEGA}{\partial
\phi_l}\right|}{\OMEGA}
$$ are uniformly bounded by $D < \infty$ and $B < \infty$
respectively, $\forall g,h\in\cG$, $u\in \cU$, $y\in \cY$, $\theta \in
\RE^{\NP}$ and $\phi\in\RE^{\NI}$. 
\end{assumption}
Parameterized distributions satisfying the final two assumption
include the softmax distribution.

We seek
$\theta \in\RE^{\NP}$ and $\phi\in\RE^{\NI}$
maximising the {\em long-term average reward}
\begin{equation}
\label{eq:1}
\eta(\theta,\phi) := \lim_{T\rightarrow\infty} 
\frac1T\E_{\theta,\phi} \left[ \sum_{t=1}^T r(i_t) \right],
\end{equation}
where $\E_{\theta,\phi}$ denotes the expectation over all trajectories
$(i_0,g_0), (i_1,g_1),\dots,$ with transitions generated according to
$P(\theta,\phi)$. Under assumption
\ref{ass:station}, $\eta(\theta,\phi)$ is independent of the starting
state $(i_0,g_0)$ and is equal to
\begin{equation}
\eta(\theta,\phi) = \sum_{i=1}^{\NS}\sum_{g=1}^{\NB} 
\pi_{i,g}(\theta,\phi) r(i) = \pi'(\theta,\phi) r,
\label{eq:eta}
\end{equation}
where $r := \left[r(1,1), \dots, r(\NS,\NB)\right]'$ and $r(i,g)$ is
defined to be $r(i)~\forall g\in\cG$.

%


\subsection{Policy-Gradient Approaches}

Policy-gradient methods do not seek to associate values with policies,
instead they adjust the policy parameters $\phi$ and $\theta$ directly
in the direction of increasing average reward, $\eta(\theta,\phi)$.
Under mild assumptions, they ensure convergence to at least a local
maximum of $\eta(\theta,\phi)$. This behaviour is in contrast to
value-function based methods that can diverge, especially when using
function approximation \cite{baird:1999}.


By adjusting the FSC parameters $\phi$, the agent learns to use
I-states to remember only what is needed to act optimally.  This
process is an automatic quantisation of the \emph{belief state} space,
providing the best policy \emph{representable by $\NB$ I-states}.  As
$\NB \rightarrow \infty$ we can represent the optimal policy
arbitrarily accurately.  Learning FSC parameters is equivalent to
searching the space of \emph{policy graphs} \cite{meuleau-II:1999}.
Policy search appeals intuitively because it is often easier to learn
\emph{how} to act than it is to learn the  \emph{value} of acting.  In many
cases it is possible for a large POMDP to be controlled well by a
relatively simple policy graph, even though the corresponding value
function is extremely complex.





\section{Model-Based Policy-Gradient}
\label{sec:model-based-policy}

In some situations an accurate model of the world may be available.
For example, manufacturing plants are often well modelled. In this
section we present an algorithm for computing estimates of the
gradient of the average reward given a model of the world. This
algorithm is feasible for many thousands of states, an improvement
over existing model-based algorithms which typically handle at most
hundreds of states\cite{geffner:1998}.

Dropping the explicit dependence on $\theta,\phi$, we can rewrite
\eqref{eq:eta} as $\eta = \pi' r$, hence $\nabla\eta = (\nabla\pi')
r$, which should be understood as $\NP + \NI$ equations, one for each
of the $\theta$ and $\phi$ parameters. Differentiating both sides of
the balance equations $\pi' P =\pi'$ (Assumption \ref{ass:station}),
yields $(\nabla\pi' )[I - P] = \pi' \nabla P$.  We condition $[I -
P]$ to make it invertible.  Let $e$ denote the $\NS\NB$-dimensional
column vector consisting of all ``$1$s'', and $e\pi' $ the
$\NS\NB\times
\NS\NB$ matrix with the stationary distribution $\pi' $ in each row.
Since $\sum_{i,g}\pi_{i,g} = 1$ we have $\sum_{i,g}\nabla\pi_{i,g} =
0$ and so $(\nabla\pi') e\pi' = 0$. Hence, we can write $(\nabla\pi')
[I - P + e\pi'] = \pi' \nabla P$. Justification for the invertibility
of $[I - P + e\pi']$ may be found in
\cite{baxter:2001}. Substituting into the expression for $\nabla\eta$ yields
\begin{equation}
\label{eq:grad2}
\nabla\eta= \pi' (\nabla P) \left[I - P + e\pi' \right]^{-1} r. 
\end{equation}

Now let $x := [I - P + e\pi' ]^{-1} r$, so that $\nabla\eta = \pi'
(\nabla P) x$.  In general, computing $x$ is $O(\NS^3 \NB^3)$;
infeasible for $\NS \NB$ greater than a few 100's of states.  To
finesse this difficulty we turn to approximate methods.  Because $[I -
P + e\pi' ]^{-1}$ is invertible we can approximate it with the series
matrix expansion $x_N =\sum_{n=0}^N \left( P^n - e\pi' \right) r$ (in
the limit $x_N$ approaches $x$), which is essentially Richardson
iteration \cite[\S 4.6]{kincaid:1991}.  Since $(\nabla P) e \pi' = 0$
we can simplify this to $x_N =\sum_{n=0}^N P^n r$. The sum can be
efficiently evaluated by iteratively computing $v_{n+1} := P v_n$,
with $v_0 := r$, and accumulating. Because this is a series of
matrix-vector multiplications, the computation has worst case
complexity $O(\NS^2 \NB^2 N)$.  $P$ is typically sparse since only a
subset of world-states $j\in\cS$ have non-zero probabilities of being
reached from some other state. For example, robots typically transit
to a small set future states regardless of $\NS\NB$. Thus practical
POMDPs exhibit complexity closer to $O(c\NS\NB N)$ where $c\ll\NS\NB$.

The remainder of the computation of $\nabla\eta$ involves computing
$\pi'\nabla P$.  Computing $\nabla P$ is in the worst case $O(\NS^2
\NB^2 (\NP + \NI) \NY \NA)$.  Sparsity again helps since $\QJIU$ and
$\NU$ are often 0, achieving practical complexities of $O(c \NS \NB
(\NP+\NI) \NA)$ where $c\ll\NS\NB\NY$.  The stationary distribution
$\pi$ is the leading left eigenvector of $P$, which is expensive to
compute exactly.  We use the power method: $\pi'_{n+1} = \pi'_n P$,
stopping when $\| \pi'_{n+1} - \pi'_n \|_\infty < \epsilon$.  The
algorithm consisting of approximating $x$ and $\pi$ by
iteration\footnote{Advanced iteration techniques such as Krylov
  subspace methods for $x$ and the Lanczos method for $\pi$ are
  worthy of further investigation.}  and computing $\widehat{\nb_N
  \eta} = \pi' (\nb P) x_N$ has been named \GAMP\ for Gradient
Approximation of Modelled POMDPs.

\subsection{Asymptotic Convergence of \GAMP}

\begin{figure}
\begin{center}
\includegraphics[height=4cm,width=6cm]{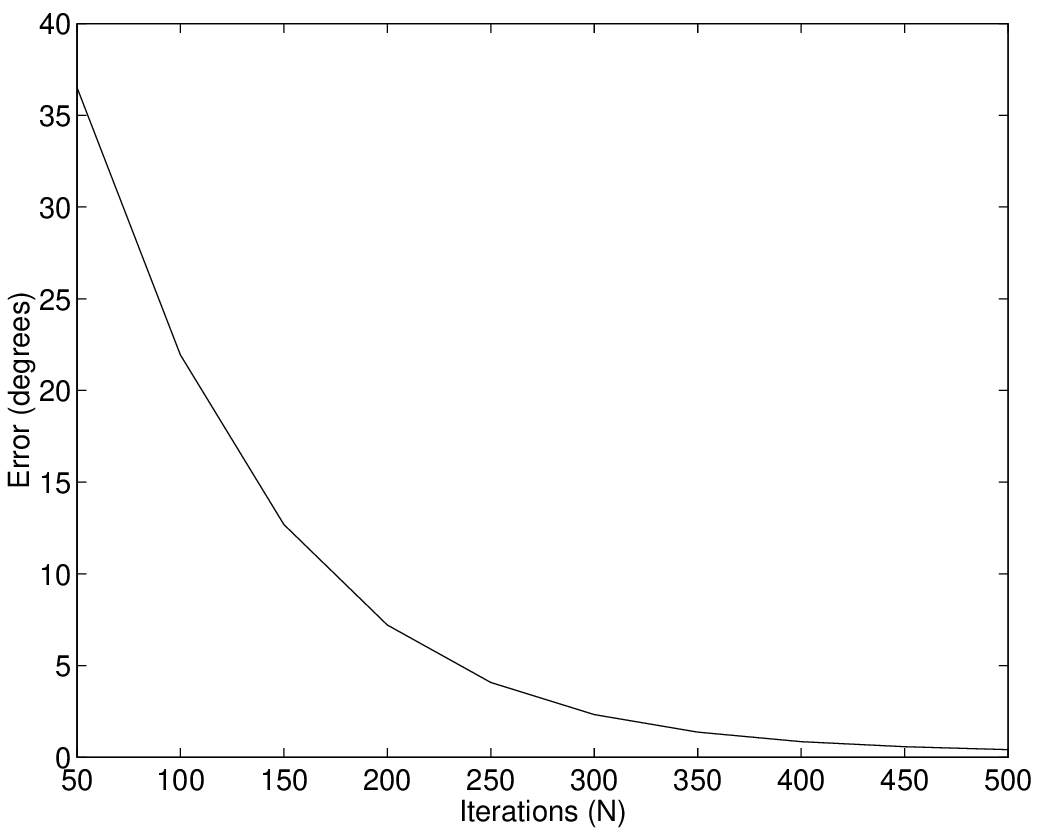}
\caption{
\label{f:GAMP-converge}
  Angular error between $\widehat{\nb_N \eta}$ after $N$ iterations
  and the exact gradient for the Pentagon problem.}
\end{center}
\end{figure}

Under assumption \ref{ass:station}, $P^N$ converges exponentially
quickly to $\pi$.  The exact rate is governed by the mixing time of
the POMDP, defined as the smallest $\tau$ which satisfies $\max_i
\sum_{j\in\cS} | P^N_{ij} - \pi_j | \le \exp(-\lfloor \frac{N}{\tau}
\rfloor)$. We can use $\tau$ to bound the estimate error after $N$ 
iterations.
\begin{theorem}
\label{theorem:asympt-conv-gamp}
$$ 
\| \widehat{\nb_N \eta} - \nb \eta \|_{\infty} 
  \le BR \NA  \tau \frac{\exp(- \lfloor \frac{N}{\tau} \rfloor)}{1 - \exp(-1)}.
$$
\end{theorem}
The difficulty of calculating $\tau$ for an arbitrary POMDP makes it
hard to use this theorem to establish $N$ \emph{a priori}. In practice
we check for convergence of $x$ every $k$ steps, stopping when
$\|x_{N+k} - x_N\|_\infty < \epsilon$.\footnote{It is easy to verify
  that $\|x_{N+1} - x_N\|$ is decreasing.}  Figure~\ref{f:GAMP-converge}
shows how quickly \GAMP\  converges on
a problem with $1045$ states. We inverted 
$[I - P + e\pi']$ exactly for the Pentagon problem defined in
Section~\ref{sec:pentagon} with $\NS=209$, $\NB=5$. Even the noisy
Pentagon transition probabilities are very sparse with only 2.4\% of
$P$ containing non-zero elements.
This experiment was run on a Pentium II @ 433 MHz. When
computing $\nb\eta$: $\pi$ takes 315 s (wall clock time), $[I - P +
e\pi']^{-1}$ takes 10.5 s and $\nabla P$ takes 36 s.  When
computing $\widehat{\nb_N \eta}$: $\pi$ takes 3.50 s
($\epsilon=0.0001$, 1319 iterations), and Richardson matrix inversion
takes 1.41 s ($N=500$).  The angular error in the gradient at $N=500$
is $0.420^\circ$ taking 11.3\% of the time the true gradient requires.
The speedup becomes greater as $\NS \NB$ grows.  Approximating $\pi$
accounts for $0.016^\circ$ of the error.

\subsection{A Multi-Agent Problem}
\label{sec:large-problem}

\begin{figure}
\begin{center}
\includegraphics[scale=0.8]{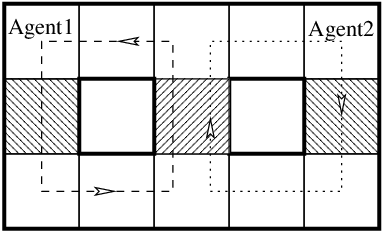}
\caption{Plan of the factory floor with 13 locations. The dashed
  lines show an optimal policy.}
\label{f:factory}
\end{center}
\end{figure}

Figure~\ref{f:factory} represents a factory floor occupied by 2 robots
which are identical except that one is given priority in situations
where both robots want to move into the same space.  The four actions
for each agent are \{{\tt{move forward, turn left, turn right,
    wait}}\}, resulting in a cross product of $\NA= 16$.  One agent
learns to move unfinished parts from the left shaded location to the
middle, where the part is processed instantly, ready for the second
agent to move the processed part from the middle to the right shaded
location. Agents only need to exit the shaded locations to load or
unload.  Each agent can be empty or carrying a part in any of 13
locations and 4 orientations ($2 \times 13 \times 4=104$ states).
The global state is the state of the 2 agents plus whether a part is
waiting at the middle, giving $104^2 \times 2 = 21,632$ states.  A
reward of 1 is received for dropping off a processed part at the right
and 0 otherwise.  To receive the maximum reward the robots must
cooperate without explicit communication.
The robots see 4 bits describing which directions are blocked by
walls, and a 5th bit describing if the agent is in the top
corridor (breaking the symmetry of the map), giving
$\NY=1024$. Uncertainty is added with a 10\% chance of the robots'
actions failing, resulting in no movement, and a 10\% chance of the
robots' sensors failing, receiving a ``no walls'' observation.  This
problem can be solved without internal state.
Section~\ref{sec:empir-comp} demonstrates \GAMP\ on problems which
require memory to solve.

These experiments were run on an AMD Athlon @ 1.3 GHz. \GAMP\ 
required less than 47 Mb of RAM to run this problem.  Unless stated,
all experiments use policies and FSCs parameterised by a table indexed
by $(y, g)$. Each index provides a set $\cM$ of $\NB$ or $\NA$ real
numbers (initialised to 0), for $\OMEGA$ and $\MU$ respectively.  The
\emph{softmax} function is then used to generate distributions 
according to $Pr(m) = \frac{\exp(m)}{\sum_{m'\in\cM} \exp(m')}$.
Computing the gradient of the softmax function with respect to $\phi$
and $\theta$ gives us $\nb \OMEGA$ and $\nb \MU$ which is needed to
compute $\nb P$. All experiments use Polak-Ribi\'{e}re
conjugate-gradient ascent \cite[\S5.5.2]{fine:1999} with a line search
(see \singleemcite{aberdeen:2002}).

The agents learn to move in opposing circles (Figure~\ref{f:factory})
around the factory,  reducing the chance of collision. They also
learn to wait when their sensors fail, using the wait action to gather
information. Table~\ref{t:factory-res} shows \GAMP\ consistently
learns the hand designed policy.

It is important to note that \GAMP\ does not use samples to estimate
gradients. This is critical in applications where random world
interactions are slow, expensive or dangerous. Fewer samples may be
needed to estimate the model parameters, prior to running \GAMP, than
to run model free algorithms.
\begin{table}
{\small
\begin{center}
\caption{Results for multi-agent factory setting POMDP over 10 runs.
$\eta$ values are multiplied by $10^{2}$.}
\begin{tabular}{l|l|l|l}
\hline
\em{Algorithm}& {\em mean $\eta$} & {\em max $\eta$} & \emph{secs to $\eta=5$} \\
\hline
\GAMP\ & 6.51 & 6.51 & 1035 \\
Hand & 6.51 & NA & NA \\
\hline
\end{tabular}
\label{t:factory-res}
\end{center}
}
\end{table}
\singleemcite{meuleau-II:1999} 
computes the gradient of the \emph{discounted} reward, avoiding
evaluation of $\pi$, but running a matrix inverse iteration for each
parameter.  Thus we perform two iterative solves to their $\NP+\NI$.


\section{Model-less Policy-Gradient}
\label{sec:model-less-policy}

Without knowledge of the POMDP transition and observation
probabilities, we resort to Monte-Carlo like methods for
estimating $\nb\eta$. These methods sample trajectories through the
POMDP, gathering information about the true gradient at each step. The
corresponding value-based algorithms, such as \TD, have been
successful for large MDPs.  We now develop the \IGPOMDP\ algorithm
which estimates $\nb \eta$ by interacting with the world. The
algorithm extends \WILLR\ with memory \cite{peshkin:1999} to the
infinite-horizon case.


\subsection{The \IGPOMDP\ Algorithm}
\label{sec:igpomdp-algorithm}


Algorithm~\ref{alg:igpomdp} computes estimates $[\Delta_T^\theta,
\Delta_T^\phi]$ of an approximation to $\nb \eta$.  At each step an
observation $y_t$ is received and $\omega$ is
evaluated to choose $g_{t+1}$. The gradient of $\OMEGAT$ is
added into the trace $z^\phi_t$ which is discounted by $\beta \in
[0,1)$ to give more weight to recent I-state choices.
The same process with the new I-state chooses an action
$u_t$.  At each step the immediate reward of action $u_t$ is
multiplied by the current traces and averaged to form the gradient
estimate. The discount factor is necessary to solve the \emph{temporal
  credit assignment problem} and reflects the assumption that rewards
are more likely to be generated by recent actions. \WILLR\ avoids 
discount factors by assuming finite-horizon POMDPs. 
The following theorem establishes that \IGPOMDP\ estimates 
\emph{an approximation} to the gradient of $\eta$ (\ref{eq:1}).
\begin{theorem}
\label{theorem:igpomdp-converge}
Define the discounted reward as 
$$
J_\beta(i,g) :=
  \E\left[\sum_{t=0}^\infty \beta^t r(i_t,g_t) | i_0 = i, g_0 = g\right], 
$$
where the expectation is over all world/I-state trajectories.
Let $\Delta_T := \left[\Delta^\theta_T, \Delta^\phi_T\right]$ be the
estimate produced by \IGPOMDP\ after $T$ iterations. Then under
Assumptions~\ref{ass:station}--\ref{ass:ratiobound},
$\lim_{T\rightarrow\infty} \Delta_T = \pi' (\nabla P) J_\beta$ w.p.1.
\end{theorem}
The next theorem establishes that as $\beta \rightarrow 1$, the
approximation $\pi' (\nabla P) J_\beta$ is $\nb \eta$.
\begin{theorem}
\label{theorem:approx}
For $P$ parameterised by $\phi$ and $\theta$
$$
\lim_{\beta\rightarrow 1} \pi' (\nabla P) J_\beta  = \nabla\eta.
$$
\end{theorem}
\begin{algorithm}[b]
\caption{\IGPOMDP}
\label{alg:igpomdp}
\begin{algorithmic}[1]
\WHILE{$t < T$}
\STATE Observe $y_t$ from the world.
\STATE Draw $g_{t+1}$ from $\omega$ and $u_t$ from $\mu$.
\STATE $z^\phi_{t+1} = \beta z^\phi_t + \frac{\nabla
\OMEGAT}{\OMEGAT}$ 
\STATE $z^\theta_{t+1} = \beta z^\theta_t + \frac{\nabla
\MUT}{\MUT}$ 
\STATE $\Delta^\theta_{t+1} = \Delta^\theta_t +
\frac1{t+1}\left[r(i_{t+1}) z^\theta_{t+1} - \Delta^\theta_t\right]$ 
\STATE $\Delta^\phi_{t+1} = \Delta^\phi_t +
\frac1{t+1}\left[r(i_{t+1}) z^\phi_{t+1} - \Delta^\phi_t\right]$ 
\STATE Issue action $u_t$.
\STATE $t++$.
\ENDWHILE
\end{algorithmic}
\end{algorithm}

 Thus $\Delta_T \xrightarrow{T\rightarrow\infty}
\pi' \nabla P J_\beta \xrightarrow{\beta\rightarrow 1} \nabla\eta$,
however the variance of $\Delta_T$ scales as
$1/\left[T(1-\beta)\right]$, reflecting the increasing difficulty of
the credit assignment problem as $\beta\rightarrow 1$. Fortunately,
$\pi' (\nabla P) J_\beta$ is guaranteed to be a good approximation to
$\nb\eta$ provided $1/(1-\beta)$ exceeds the mixing time $\tau$.  The
proofs are a generalisation from the memoryless \GPOMDP\ algorithm
\cite{baxter:2001} which is retrieved from \IGPOMDP\ by setting
$\NB=1$.

\subsection{Zero Gradient Regions}
\label{sec:zero-grad-regi}
\label{sec:sparse-fscs}

We observed that FSC methods with \emph{small random initial
  parameters} fail to learn to use the FSC for non-trivial problems.
In this case $\OMEGA$ and $\MU$ are near uniform and therefore the
distribution over I-state trajectories, given different observation
histories, is also near uniform.  Thus, it is difficult for the
controller to distinguish high reward I-state trajectories and which
I-state transitions to adjust in order to increase reward. This
results in near zero gradient with respect to the internal-state
parameters.  The following theorem formalises this argument.
\begin{theorem}
\label{sec:zero-grad-regi-1}
If we choose $\theta$ and $\phi$ such that $\OMEGA = \omega(h|\phi,
g', y)~\forall g,g',h,y$ and $\MU = \mu(u|\theta, h', y)~\forall
h,h',y,u$ then $\nabla^\phi \eta = 0$.
\end{theorem}
The conditions of the theorem may be violated by one iteration of
gradient ascent. \singleemcite{aberdeen:2002} establishes the
slightly more restrictive conditions which result in \emph{perpetual}
failure to learn a finite state controller. We have investigated
avoiding small gradient regions by using a large \emph{sparse}
I-state FSC where all I-states have out-degree $k\ll \NB$. This trick
ensures that distinct observation histories typically generate
minimally overlapping distributions of I-state trajectories.  The
complexity of each \IGPOMDP\ step grows linearly with $k$, not $\NB$,
allowing the use of many I-states to compensate for the loss in FSC
richness.  Section~\ref{sec:heavenhell} contains an empirical
demonstration of the effect of sparse transitions.


\section{Internal Belief States}
\label{sec:intern-beli-stat}

\IGPOMDP\ discards information because it uses Monte-Carlo sampling to 
estimate the effect of internal-state transitions, even though the
internal-state transition probabilities are known to the algorithm
(they are given by $\omega$). This results in an unnecessary increase
in the variance of the algorithm's gradient estimates. To avoid this,
at each time step we can update the probability of occupying each
I-state and use this \emph{belief over I-states} as the agent's
memory. Our first attempt in this direction used Input/Output HMMs
\cite{bengio:1996}, computing hidden state occupancy probabilities
driven by the observations.  The problem with existing IOHMM/POMDP
algorithms such as \singleemcite{chrisman:1992}, is that they ignore
the most useful indicator of performance: the reward. Our initial
approach used the IOHMM to \emph{predict rewards}. However, this
suffers from the same difficulties as value-function methods: the
agent can learn to predict rewards well while failing to learn a good
policy \cite{aberdeen:2002}.

\EGPOMDP\ is an alternative which overcomes the limitations of
predicting rewards.
The algorithm is a partly Rao-Blackwellised version of
\IGPOMDP; computing the expectation only over I-state trajectories.
Rao-Blackwellisation of Monte-Carlo estimators reduces their variance
\cite{casella:1996}.

We replace $\mu$ with $\Bmu$ which implicitly contains the I-state in
the form of a belief over I-state occupation probabilities.  The
I-state update is still parameterised by a stochastic FSC described by
$\phi$, but the update to the internal \emph{belief} state is not
stochastic.  Let $\alpha_t(g|\phi, y_t, \alpha_{t-1})$ be the
probability that $g_t = g$ given the current parameters, observation
and the previous I-state belief. The recursive I-state update is
\begin{equation}
\label{eq:21}
\alpha_{t+1}(h | \phi, y, \alpha_t) = 
   \sum_{g\in\cG} \alpha_t(g | \phi, y, \alpha_{t-1}) \OMEGA.
\end{equation}
If the initial world state of the system is deterministic then we make
$\alpha_0$ deterministic, otherwise we can set
$\alpha_0(g)=1/\NB,~\forall g$.  The new form of the
policy which takes the expectation of $\mu$ over internal states is
\begin{equation}
\label{eq:31}
\Bmu(u | \theta, \phi, y) := \sum_{g\in\cG} \alpha_{t+1}(g | \phi, y, \alpha_t) \mu(u|\theta,g,y).
\end{equation}
We compute $\nabla \Bmu$ by application of the chain-rule
to~\eqref{eq:31}, where $\nabla \alpha_{t+1}$ is computed recursively
from $\nabla \alpha_t$, shown in Algorithm~\ref{alg:egpomdp}.  Each
gradient estimation step has a complexity of at least $O(\NB(\NB +
\NA))$ compared to the \IGPOMDP\ complexity of $O(\NB + \NA)$ per
step.  This becomes $O(\NB(k + \NA))$ using the sparse transition
trick of Section~\ref{sec:sparse-fscs}, also overcoming the small
gradient problem.  \EGPOMDP\ has the same convergence properties as
\IGPOMDP.  \begin{algorithm}[t]
\caption{\EGPOMDP}
\label{alg:egpomdp}
\begin{algorithmic}[1]
\raggedright
\STATE Set $z_0 = 0$, and $\Delta = 0$ ($z_0, \Delta_0 \in\RE^{\NP + \NI}$).
\FOR{each $g\in\cG$}
\STATE Set $\alpha_0(g) = 1/\NB$ and $\nabla \alpha_0(g) = 0$.
\ENDFOR
\WHILE{$t < T$}
\STATE Observe $y_t$ from the world.
\FOR{each $h\in{\cG}$}
\STATE $\alpha_{t+1}(h | \phi, \alpha_{t}) = 
        \sum_{g\in\cG} \alpha_t(g | \phi, \alpha_{t-1}) \omega(h | \phi, g, y_t).$
\STATE  $\nb \alpha_{t+1}(h | \phi, \alpha_{t}) = 
        \sum_{g\in\cG}
        \nb \alpha_t(g | \phi, \alpha_{t-1})\omega(h | \phi, g, y_t) +
        \alpha_t(g | \phi, \alpha_{t-1})\nb \omega(h | \phi, g, y_t).$
\ENDFOR
\STATE Choose $u_t$ from $\Bmu(u_t | \theta, \phi, y_t) = 
       \sum_{g\in\cG} \alpha_{t+1}(g | \phi, \alpha_t) \mu(u|\theta,g,y_t)$.
\STATE $z_{t+1} = \beta z_t + \frac{\nb
\Bmu(u_t | \theta, \phi, y_t)}{\Bmu(u_t|\theta, \phi, y_t)}$ 
\STATE $\Delta_{t+1} = \Delta_t +
\frac1{t+1}\left[r(i_{t+1}) z_{t+1} - \Delta_t\right]$ 
\STATE Issue action $u_t$.
\STATE $t++$.
\ENDWHILE
\end{algorithmic}
\end{algorithm}


\EGPOMDP\ is similar to the finite-horizon algorithm presented by
\singleemcite{shelton:2001}, an HMM gradient ascent algorithm where
emissions are actions. In that paper the HMM backward probability is
used as well as the forward probability $\alpha$. In the
infinite-horizon setting there is no natural point to begin the
backward probability calculation so our algorithm uses only $\alpha$.


\section{Empirical comparisons}
\label{sec:empir-comp}
\subsection{Heaven/Hell}
\label{sec:heavenhell}
\begin{figure}[t]
\begin{center}
\includegraphics[width=7cm,height=2.5cm]{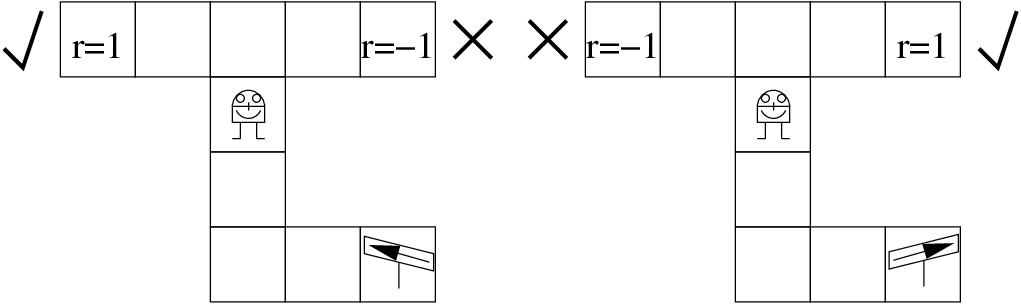}
\caption{The Heaven/Hell problem: the optimal policy is to visit the
  signpost to find out whether to move left or right at the top-middle
  state in order to reach heaven.}
\label{f:hh}
\end{center}
\end{figure}

The Heaven-Hell problem of \singleemcite{geffner:1998} is shown in
Figure~\ref{f:hh}. The agent starts in either position shown with
probability 0.5. The state is completely observable except that the
agent does not initially know if it is in the left world or the right
world. The agent must first visit the signpost which provides this
information, and remember the sign direction until the top-middle
state. In theory, 3 I-states are sufficient for optimal control.  Two
features make this problem hard: multi-stage memory is needed,
and temporal credit assignment is difficult because the rewards
are delayed up to 11 steps from relevant actions.  We used lookup
tables with softmax distributions yielding a total of 1540 parameters.
The gradient estimation time in \IGPOMDP\ was set to $T=10^7$, with
$\beta = 0.99$. We chose 20 internal states for the controller,
yielding a total of $\NS \NB=400$.

To demonstrate the necessity of sparse I-state transitions
(Section~\ref{sec:zero-grad-regi}) the experiments were run with fully
connected FSCs and randomly connected FSCs with an outdegree of
$k=3$. The randomness in the \GAMP\ results comes only from the choice
of initial FSC.  Table~\ref{t:hh-results} shows the results of these
experiments.  Both algorithms find the optimum policy, but only \GAMP\
finds it consistently.  The experiments with dense FSCs failed with
\GAMP\ producing gradient estimates within machine tolerance of
0.
\begin{table}[t]
{\small
\begin{center}
\caption{Results
  for the Heaven/Hell scenario over 10 runs.  $\eta$ values are
  multiplied by $10^{2}$. Bracketed numbers indicate how many runs
  converged to $\eta=5.0$.}
\begin{tabular}{l|l|l|l|r}
\hline
{\em Alg} & {\em ave} & {\em max}& {\em var} & {\em secs to $\eta=5$} \\
\hline
\GAMP\ $k=3$      & 9.01   & 9.09  &0.0514& 34 (10) \\
\GAMP\ dense      & 0.005 & 0.005 & $<.001$&  (~0) \\
\sf{IState} $k=3$ & 6.49   & 9.09   & 10.8& 11436 (~8) \\
\sf{IState} dense & 0.018 & 0.034 &$<.001$& (~0)\\
\hline
Optimum & \multicolumn{4}{l}{9.09 (11 steps between rewards)}\\
\hline
\end{tabular}
\label{t:hh-results}
\end{center}
}
\end{table}

The wall clock times to convergence quoted for these experiments are
not directly comparable. \IGPOMDP\ was run using 94 processors of a
550 MHz dual CPU PIII Beowulf cluster. \GAMP\ was run on a 1.3 GHz
Athlon, roughly equivalent to 3 CPUs of the cluster. This demonstrates
the advantage of having a model of the world.  Within our knowledge
this is the first time the Heaven/Hell problem has been solved using a
model-less algorithm.\footnote{We verified that \EGPOMDP\ can learn to visit
the signpost. However, since 1 run takes $>2$ days on our cluster,
we were prohibited from a 
formal comparison.}

\subsection{Pentagon}
\label{sec:pentagon}
\begin{figure}
\begin{center}
\includegraphics[height=7cm,width=4.2cm,angle=270]{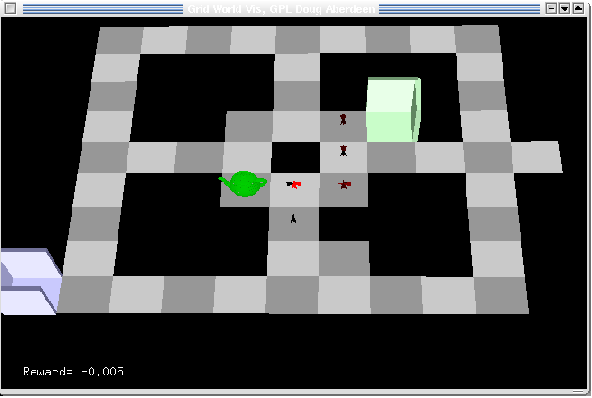}
\caption{The Pentagon robot navigation problem maze.}
\label{fig:pentagon}
\end{center}
\end{figure}

In this problem a robot must navigate through corridors to reach a goal
(Figure~\ref{fig:pentagon}).  The world is mapped into discrete
locations. The robot can point North, South, East or West. The actions are \{\texttt{move forward}, \texttt{turn left}, \texttt{turn right},
\texttt{declare goal}\} and they fail with 12\% probability. There are
28 observations indicating if the locations in front and to the sides
of the robot are reachable. Observations have a 12\% chance of being
wrong.  We modified the
original POMDP definition in \cite{cassandra:1998} to make rewards
control independent and to make the agent jump to the start state
when it achieves the goal. No penalty is received for 
incorrect \texttt{declare goal} actions. The agent 
starts in the same place but noise means the agent quickly
becomes confused about its location even with perfect memory of
observations/actions.
We have $\NS=209$ and  $\NB\le20$, giving at most 4180 states. The
\IGPOMDP\ algorithm required a minimum of $2 \times 10^6$ gradient
estimation steps, while the \EGPOMDP\ algorithm required $10^6$
steps. For comparison purposes we have reported results for
\BELIEF, a variation on \IGPOMDP\ in which the observations $y$ are
replaced with the entire belief-state vector over
world-states. Actions in \BELIEF\ are generated by a softmax distribution
that computes a linear combination of the world-state occupation
probabilities for each possible action. Since the belief-state vector
contains all information about observation histories, \BELIEF\ does not
maintain any internal state.

\begin{table}
{\small
\begin{center}
\caption{Results on the Pentagon scenario over 10 runs.
  $\eta$ values are multiplied by $10^{2}$.}
\begin{tabular}{l|l|l|l|l|l|r}
\hline
\emph{Alg} & \emph{$\NB$} & \emph{k} &\emph{ave} & \emph{max} & {\em var} & {\em secs to $2$}\\
\hline
\GAMP\     &5&2  & 2.55 & 2.70 &0.0102 & 611 (10)     \\
           &10&2 & 2.50 & 2.63 &0.0128 & 4367 (10)    \\
           &20&2 & 2.50 & 2.80 &0.0250 & 31311 (10)   \\
           &20&3 & 2.89 & 3.00 &0.00613 & 47206 (10) \\
\sf{IState} &5 &2 & 2.06 &2.42 &0.281 & 649 (~9)   \\
           &10&2 & 2.18 & 2.37 &0.0180 & 869 (~9)   \\  
           &20&2 & 2.12 & 2.28 &0.0137 & 1390 (~9)  \\
           &20&3 & 2.15 & 2.33 &0.0138 & 1624 (~9)  \\
\sf{Exp}  &5 &2 & 1.96 & 2.33 & 0.401 &  1708 (~7)  \\
           &10&2 & 2.19 & 2.40 &0.0151 & 6020 (~9)  \\
           &20&2 & 2.11 & 2.27 &0.0105 & 29640 (~8)  \\
           &20&3 & 2.26 & 2.36 &0.00448 & 48296 (10)  \\
\hline
\sf{IState}     & 1& 1& 1.35 & 1.37 & 0.00032 & (~0)      \\
\BELIEF\          &  &  & 2.67 & 3.65 &0.778 & 2313 (~7) \\
MDP             &  &  & 4.93 & 5.01 &  0.00148& 24  (10) \\
\hline
\end{tabular}
\label{t:pentagon-results}
\end{center}
}
\end{table}

I-states approximate the policies achievable using a full belief state
thus the \BELIEF\ result of $\eta=3.65$
(Table~\ref{t:pentagon-results}) is an approximate upper bound for the
$\eta$ we should be able to achieve using a large number of I-states.
\IGPOMDP\ with $\NB=1$ is a memoryless policy which should lower bound
the I-state results. Values of $\eta$ between these empirical bounds
show that a finite number of I-states can be used to learn better than
memoryless agents.

We ran our trained agents on the original POMDP to compare results
with \singleemcite{cassandra:1998} which uses value-iteration and the
most likely state heuristic.  Our best \BELIEF\ agent ($\eta=3.65$)
achieves a discounted reward of 0.764, requiring an average of 27
steps to achieve the goal.  This result is bracketed by Cassandra's
results of 0.791 (known start state) and 0.729 (uniform initial belief
state). 

Because the \BELIEF\ algorithm has access to the world-state belief
(using the model), it is unsurprising that it obtains the best maximum
$\eta$.  \GAMP\ also uses a model but restricts memory to $\NG$
states, resulting in the second highest maximum.  \GAMP\ may sometimes
be preferable to \BELIEF\ because of its zero-variance (the variance
in the tables is due to the random choice of sparse FSC), supported
by the fact that for $\NB=20$ the \GAMP\ \emph{mean} is better than
\BELIEF.

\begin{figure}
\begin{center}
\includegraphics[width=73mm,height=55mm]{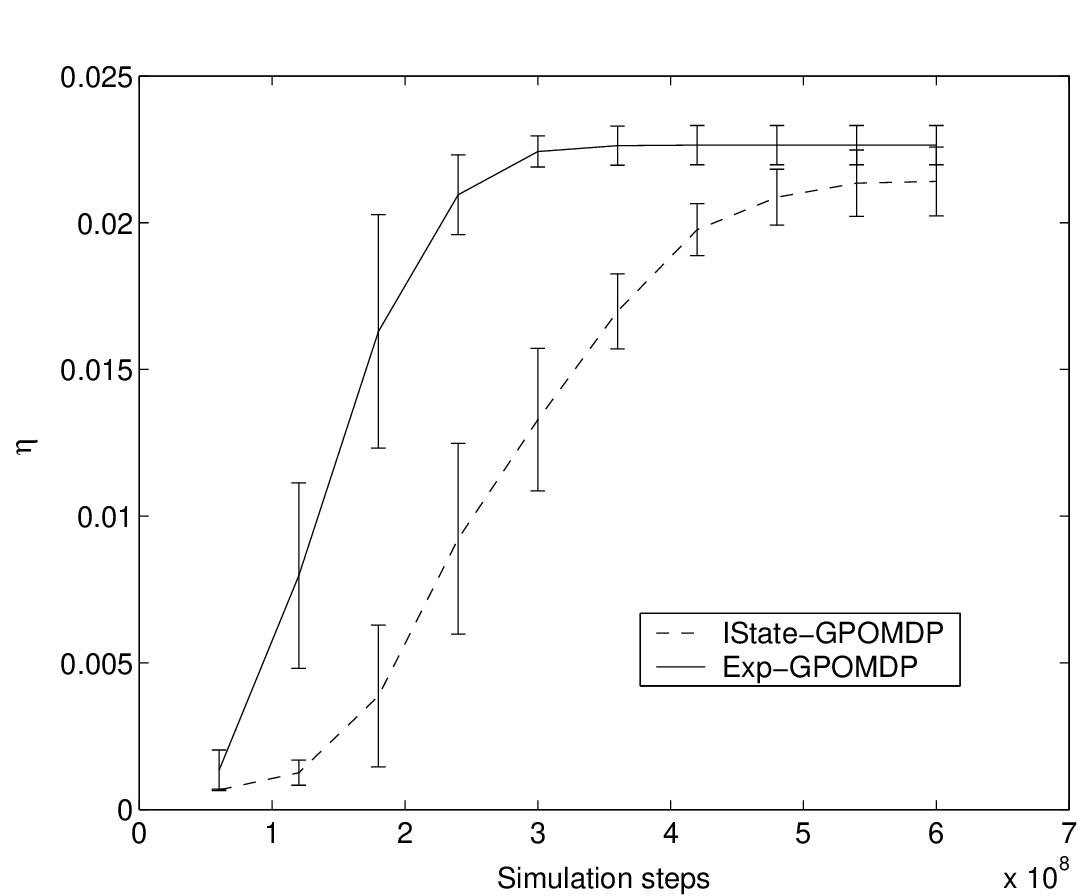}
\caption{
  Convergence of \IGPOMDP\ \emph{vs.} \EGPOMDP\ for 10 Pentagon runs with
  $\NB=20$ and $k=3$. 
\label{fig:pentagon-converge}}
\end{center}
\end{figure}

Increasing $\NB$ and the FSC connectivity $k$ generally improves the
results due to the increasing richness of the parameterisation,
however, \IGPOMDP\ performed best for $\NB < 20$ because we did not
scale the number of gradient estimation steps with $\NB$.  {\EGPOMDP}s
reduced variance allows it to improve consistently as $\NB$ is
increased while still using fewer estimation steps than \IGPOMDP.
Figure~\ref{fig:pentagon-converge} shows that for $\NB=20$, \EGPOMDP\ 
produces a superior result in fewer steps than \IGPOMDP.  Setting $k >
3$ caused failures due to small initial gradients.

\section{Conclusion}

Memoryless algorithms are inadequate for all but trivial POMDPs.
However, even for approximate algorithms, methods that use the belief
state become intractable for large $\NS$. We have proposed and
compared 3 finite state controller algorithms which lie between
memoryless algorithms and belief state algorithms.  \GAMP\ takes full
advantage of a model to compute zero variance, low bias estimates of
the performance gradient. Use of iterative approximations, sparse
structures, and sparse FSCs allows \GAMP\ to solve infinite-horizon
POMDPs an order of magnitude larger than competing methods on a basic
desktop computer. \IGPOMDP\ and \EGPOMDP\ do not need a model and
scale in two senses: the running time is constant per step and the
number of simulation steps required does not necessarily increase with
$\NS$.  Future work includes developing factored-state
policy-gradient algorithms and incorporating further variance
reduction methods.



{\small
\bibliography{papers}
\bibliographystyle{mlapa}
}
\end{document}